\documentclass[11pt]{article}

\usepackage[final]{acl}

\usepackage{amsmath}
\usepackage{mathtools}
\usepackage{makecell}
\usepackage{marvosym}
\usepackage{amssymb}
\usepackage{pifont}
\usepackage[table]{xcolor}
\usepackage{times}
\usepackage{latexsym}
\usepackage{multirow}
\usepackage{xcolor}
\usepackage{tcolorbox}
\usepackage[T1]{fontenc}

\usepackage[utf8]{inputenc}

\usepackage{microtype}

\usepackage{inconsolata}
\usepackage{hhline}
\usepackage{graphicx}
\usepackage{listings}

\lstdefinestyle{promptbox}{
  basicstyle=\ttfamily\small,
  breaklines=true,
  columns=fullflexible,
  backgroundcolor=\color{gray!10},
  frame=single,
  rulecolor=\color{gray!70},
  showstringspaces=false
  basewidth=0.5em,
  breakindent=0pt,
  xleftmargin=0pt,
  framexleftmargin=0pt
}

\lstdefinestyle{requirements}{
  basicstyle=\ttfamily\small,
  breaklines=true,
  columns=fullflexible,
  backgroundcolor=\color{blue!5},
  frame=single,
  rulecolor=\color{gray!70},
  showstringspaces=false
  basewidth=0.5em,
  breakindent=0pt,
  xleftmargin=0pt,
  framexleftmargin=0pt
}

\title{PersonalAlign: Hierarchical Implicit Intent Alignment for Personalized GUI Agent with Long-Term User-Centric Records}

\author{
\textbf{Yibo Lyu}\textsuperscript{1},
\textbf{Gongwei Chen}\textsuperscript{1},
\textbf{Rui Shao}\textsuperscript{1,2\textdagger},
\textbf{Weili Guan}\textsuperscript{1},
\textbf{Liqiang Nie}\textsuperscript{1\textdagger} \\
\textsuperscript{1} Harbin Institute of Technology, Shenzhen \quad
\textsuperscript{2} Shenzhen Loop Area Institute
\\
weberlv1b@gmail.com \hspace{1em} \{shaorui, nieliqiang\}@hit.edu.com
\\
  \url{https://github.com/iLearn-Lab/ACL26-PersonalAlign}
}

\begin{document}
\maketitle
\begin{abstract}
While GUI agents have shown strong performance under explicit and completion instructions, real-world deployment requires aligning with users’ more complex implicit intents. In this work, we highlight Hierarchical Implicit Intent Alignment for Personalized GUI Agent (\textbf{PersonalAlign}), a new agent task that requires agents to leverage long-term user records as persistent context to resolve omitted preferences in vague instructions and anticipate latent routines by user state for proactive assistance.
To facilitate this study, we introduce \textbf{AndroidIntent}, a benchmark designed to evaluate agents’ ability in resolving vague instructions and providing proactive suggestions through reasoning over long-term user records. We annotated 775 user-specific preferences and 215 routines from 20k long-term records across different users for evaluation.
Furthermore, we introduce Hierarchical Intent Memory Agent (\textbf{HIM-Agent}), which maintains a continuously updating personal memory and hierarchically organizes user preferences and routines for personalization.
Finally, we evaluate a range of GUI agents on AndroidIntent, including GPT-5, Qwen3-VL, and UI-TARS, further results show that HIM-Agent significantly improves both execution and proactive performance by 15.7\% and 7.3\%.
\end{abstract}
{\let\thefootnote\relax\footnotetext{\textsuperscript{\textdagger} Corresponding authors.}

\section{Introduction}
With the rapid advancement of multimodal large language models (MLLM) \cite{bai2025qwen2, an2025llava}, GUI agents have made significant progress in grounding natural language instructions into executable actions \cite{wang2025ui, wu_os-atlas_2024, chen2025SimpAgent}. However, most existing works are evaluated in simulated environments and rely on the strong assumption of complete user instructions. We argue that instructions often fail to fully capture user's true intent in daily usage, motivating the need for personalized agents capable of perceiving intent beyond the traditional instruction-following reactive paradigm.

\begin{figure}[t]
  \centering
   \includegraphics[width=1.0\linewidth]{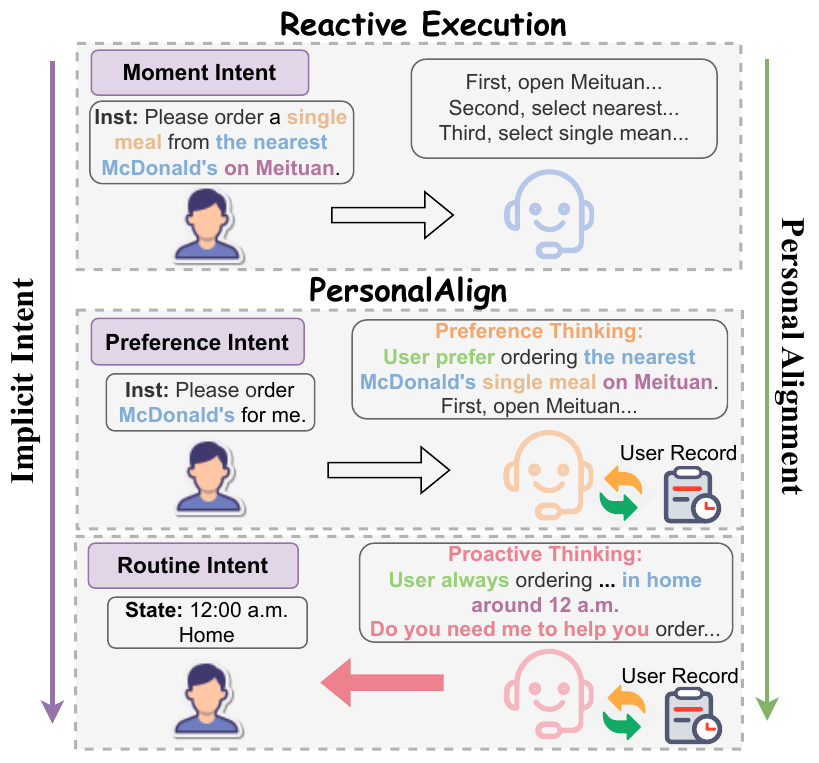}
    \caption{We highlight a new PersonalAlign agent task. Agent should leverage the user’s long-term record to provide hierarchical implicit intent alignment for both preference and routine intent.}
   \label{fig:intro}
\end{figure}

\begin{table*}[t]
\setlength{\tabcolsep}{2pt}
\footnotesize
\centering
\caption{Comparison of the AndroidIntent with existing GUI benchmark or dataset. We also show that whether each trait is fully incorporated (\textcolor{green}{\Large \ding{51}}), partially incorporated (\textcolor{yellow!70!black}{\Large \ding{51}}), or not incorporated (\textcolor{red}{\Large \ding{55}}).}
\label{tab:bench}
\begin{tabular}{c | c c c c c | c}
\hline
\makecell{\bf Benchmark or Dataset} & 
\makecell{\bf vague \\ \bf Instruction} & 
\makecell{\bf Proactive \\ \bf Suggestion} & 
\makecell{\bf Long-term \\ \bf Records} & 
\makecell{\bf User-Centric \\ \bf Annotation} &
\makecell{\bf User \\ \bf Modeling} & 
\makecell{\bf Task Target} \\
\hline
AITW \cite{rawles2023androidinthewild} & \textcolor{red}{\Large \ding{55}} & \textcolor{red}{\Large \ding{55}} & \textcolor{red}{\Large \ding{55}}& \textcolor{red}{\Large \ding{55}} & \textcolor{red}{\Large \ding{55}} & GUI Execution \\
AndroidControl\cite{li2024effects} & \textcolor{red}{\Large \ding{55}} & \textcolor{red}{\Large \ding{55}} & \textcolor{red}{\Large \ding{55}} & \textcolor{red}{\Large \ding{55}} & \textcolor{red}{\Large \ding{55}} & GUI Execution \\
SPA-Bench\cite{chenspa} & \textcolor{red}{\Large \ding{55}} & \textcolor{red}{\Large \ding{55}} & \textcolor{red}{\Large \ding{55}} & \textcolor{red}{\Large \ding{55}} & \textcolor{red}{\Large \ding{55}} & GUI Execution \\
\hline
ProactiveAgent\cite{luproactive} & \textcolor{red}{\Large \ding{55}} & \textcolor{green}{\Large \ding{51}}  & \textcolor{red}{\Large \ding{55}} & \textcolor{red}{\Large \ding{55}} & \textcolor{red}{\Large \ding{55}} & Proactive Agent \\
OS-Karois\cite{cheng2025kairos} & \textcolor{yellow!70!black}{\Large \ding{51}} & \textcolor{red}{\Large \ding{55}} & \textcolor{red}{\Large \ding{55}} & \textcolor{red}{\Large \ding{55}} & \textcolor{red}{\Large \ding{55}} & Human Cooperation \\
IFRAgent\cite{wu2025quick} & \textcolor{yellow!70!black}{\Large \ding{51}} & \textcolor{red}{\Large \ding{55}}  & \textcolor{yellow!70!black}{\Large \ding{51}} & \textcolor{red}{\Large \ding{55}} & \textcolor{green}{\Large \ding{51}} & Personalized Execution \\
FingerTip\cite{yang2025fingertip}  & \textcolor{red}{\Large \ding{55}} & \textcolor{yellow!70!black}{\Large \ding{51}} & \textcolor{green}{\Large \ding{51}} & \textcolor{red}{\Large \ding{55}} & \textcolor{red}{\Large \ding{55}} & Behavior Prediction \\
\hline
AndroidIntent & \textcolor{green}{\Large \ding{51}} & \textcolor{green}{\Large \ding{51}} & \textcolor{green}{\Large \ding{51}} & \textcolor{green}{\Large \ding{51}} & \textcolor{green}{\Large \ding{51}} & Intent Alignment  \\
\hline
\end{tabular}
\end{table*}

From a user-centric perspective, human–agent interaction constitutes a \textbf{\textit{joint activity}} where meaning is co-constructed through shared context \cite{clark1996using}.  In daily usage, this shared context leads users to naturally omit recurring patterns, assuming the agent can "fill in the blanks" from historical records \cite{cheng2025navi, qian2024tell}, leading to the emergence of \textbf{implicit intent}.
As shown in Figure \ref{fig:intro}, user intent exhibits hierarchical degrees of implicitness. While reactive agents primarily handle explicit instructions, a personalized agent should extend to align user's implicit intent by leveraging long-term user records as context: addressing \textbf{preference intent in vague instructions} where preference details are omitted, and further anticipating \textbf{routine intent without instruction} only based on current user state. Bridging implicit intent gaps is essential for effective joint activity and for establishing stronger human–agent trust.

To address this practical dilemma, we highlight Hierarchical Implicit Intent Alignment for Personalized GUI Agent (\textbf{PersonalAlign}). This task shifts focus from simple execution to align with user's implicit intent.
Specifically, PersonalAlign requires agent to identify preferences from past repeated selections to resolve vague instructions, while separating more frequent and state-consistent routines to facilitate proactive assistance. This enables agents to transition from independent reactive executors into personalized partners that co-evolve through personal interaction.
To achieve PersonalAlign, current research faces two limitations:

\textbf{1. Lack of long-term, user-centric annotated benchmarks.}
Most existing datasets primarily focus on static, simulated completion instruction execution, failing to evaluate personalized agent. To bridge this gap, we introduce \textbf{AndroidIntent}, a novel benchmark constructed from personal daily records.
We explicitly annotate user preferences and routines from long-term records, and simulate implicit intent by carefully removing recoverable personal preferences from original intents.
To mitigate subjectivity and conceptual ambiguity during the annotation process, we introduce a hierarchical filtering-verifying strategy. This approach translates abstract personalization concepts into quantifiable scores, allowing us to efficiently identify candidates for human verification. As detailed in Table~\ref{tab:bench}, AndroidIntent evaluates GUI agent’s ability to align implicit intent by leveraging long-term user records as context for user modeling.

\textbf{2. Inability to manage long-term user records.} Existing agent memory for LLM-chat often rely solely on semantic similarity, which is insufficient for handling GUI execution. Moreover, such memory fail to support hierarchical intent alignment. To address these limitations, we propose a specially designed agent memory, Hierarchical Intent Memory Agent (\textbf{HIM-Agent}).
HIM-Agent incorporates a Streaming Aggregation Module to enable incremental updates of user behavior. Building upon this foundation, the Execution-based Preference Filter and the State-based Routine Filter extract Preference Intent Memory and Routine Intent Memory, forming a hierarchical intent alignment support. 

In the experiment, we observe that vague instructions typically convey coarse-grained sub-goals, causing fine-grained execution failures when specific personal requirements are absent at certain steps, and that most current GUI agents fail to deliver reliable proactive suggestions, calling for personal agents' stronger context analysis capabilities.
Our contribution can be summarized as:

\begin{itemize}
    \item We introduce a new agent task \textbf{PersonalAlign}, which requires agent to align with user's implicit intents through long-term record.
    \item We construct a new GUI bench \textbf{AndroidIntent}, which annotates daily preference and routine intent from long-term records.
    \item We propose a new agent memory \textbf{HIM-Agent}, which hierarchically learns preference and routine intent from long-term records.
    \item We conduct extensive experiments and analysis across various GUI Agents and demonstrate superior performance of HIM-Agent.
\end{itemize}

\section{Related Work}
\label{sec:related}

\subsection{Personalized GUI Agent}
Agents assist users in operating mobile devices or interacting with the real world through natural-language instructions \cite{zhu2026delta, li2025cogvla, chen2025SimpAgent, zhu2026H-GAR}. While many datasets and methods concentrate on evaluating execution success rate \cite{rawles2023androidinthewild, lu2025guiodyssey, xie2024osworld}, a more intelligent personal GUI assistant should infer and match user’s true intent.
Some prior work focuses on learning from human action for personalization. For example, IFRAgent \cite{wu2025quick} extracts both explicit and implicit intention flows from short-term instruction traces. FingerTip \cite{yang2025fingertip} uses recent historical actions to personalize current execution.
On the other hand, some research emphasizes proactive suggestions for personalization. ProactiveAgent \cite{luproactive} trains agents to monitor the environment and user activities to enable proactive behaviors; ContextAgent \cite{yang2025contextagent} leverages real-world sensors to infer users’ current states; and Galaxy \cite{bao2025galaxy} introduces the Cognition Forest framework for proactive services.
However, existing methods rarely address daily, user-centric personalization; preference execution and proactive suggestion have largely evolved as isolated paradigms. In this work, we bridge this gap by introducing a hierarchical view of user's implicit intent, thereby unifying these two tasks into a coherent and unified, user-centric personalization framework for GUI agents.

\subsection{Agent Memory}
Memory is essential for agents to persist user long-term agent interactions \cite{bao2025galaxy, li2024optimus, long2025seeing}, which is critical for building an intelligent personal GUI agent. MemAgent \cite{yu2025memagent} and Mem1 \cite{zhou2025mem1} propose progressively compressing dialogues to extend context length, while Memory-R1 \cite{yan2025memory} introduces an RL framework to train agents in organizing and utilizing memory. Mirage-1 \cite{xie2025mirage} constructs an Execution Skill Memory to enhance GUI agent performance.
For user profiling and memory construction, Memory OS \cite{kang2025memory} aggregates user dialogues into segments and applies memory into short-, mid-, and long-term modules for retrieval. LettinGo \cite{wang2025lettingo} uses LLM to generate diverse and adaptive user profiles. PersonaX \cite{shi2025personax} clusters historical records for user profile modeling.
To address PersonalAlign, agent memory should not only generalize over long-term user records but also hierarchically organize user intents.

\section{PersonalAlign Task Definition}
For each user's long-term record, we organize the data chronologically and use the first 80\% record as historical record $\mathcal{H}=\{R_1, R_2 \dots R_m\}$, while the remaining 20\% record serves as the executing record $E=\{R_{m+1} \dots R_{n}\}$ for annotation and inference. Each GUI interaction record $R_i$ consists of the following elements: instruction $I$, interaction time $T$, interaction scenario $S$, execution action trajectory $A = [\tau_1, \tau_2, \dots, \tau_{n}]$ and screen observation $O = [o_1, o_2, \dots, o_{n}]$, formally represented as: $R_i = \{(I_i, T_i, S_i, A_i, O_i)\}_i^{N}$.

At the most fundamental level, when the user provides a complete and explicit instruction $I_t$, the agent operates under reactive GUI execution paradigm. In this setting, the agent executes a sequence of actions $\tau_t$ to fulfill the instruction, lacking the capability to leverage long-term historical records and perform personalization.

Beyond the reactive level, user's daily instructions may lack explicit preference information. In these cases, a personalized agent should infer the missing preferences based on user’s historical record $\mathcal{H}$. Given a vague instruction $\hat{I}_t$, agent should infer user’s intent from the historical intent and action sequences, ensuring each step selection satisfies the user's intent.

At a higher level, user may exhibit latent routines, requiring agents to incorporate current user state to provide proactive suggestions. By leveraging repetitive intents in historical records $\mathcal{H}$ under similar states, agents can proactively generate suggestions $I_t'$, even without user instructions, denoted as $I_\emptyset$. We formalize these three paradigms as:
\begin{equation}
\small
\begin{cases}
A_t \leftarrow f_{\theta}(I_t;(\{\}) \in \mathcal{H}) \\
A_t \leftarrow f_{\theta}(\hat{I}_t; (\{I_i, A_i\}) \in \mathcal{H}) \\[2pt]
I_t’ \leftarrow f_{\theta}(I_\emptyset; (\{I_i, T_i, S_i\}) \in \mathcal{H}),
\end{cases}
\end{equation}
where the three formulations correspond to Reactive, Preference, and Routine Intent alignment, respectively. The left side represents user inputs, while the right side denotes the historical interaction records $\mathcal{H}$ that the agent leverages. For PersonalAlign, we primarily focus on the latter two settings, where the agent should learn to apply the user's long-term records as context to provide hierarchical personalized services to improve user trust and satisfaction in daily use. More discussions on diverse forms of personalized agents and their settings are provided in Appendix \ref{sec:personal-setting}.

\section{AndroidIntent}
To evaluate PersonalAlign, we need to annotate the ground truth from continuous user records, specifically identifying user preference and routine intents that are sufficiently supported by historical evidence to enable reliable personalization. To bridge this gap, we present AndroidIntent, built upon 91 users' 2 months of Android interaction records from Fingertip20K \cite{yang2025fingertip}. However, directly annotating such 20k long-horizon user histories introduces significant challenges: the definition of preference and routine is ambiguous and lacks objective standards. To alleviate these issues and ensure annotation quality, we introduce a new hierarchical filtering-verifying strategy. As shown in Figure~\ref{fig:pipeline}, we illustrate the annotation pipeline.

\subsection{Hierarchical Filtering Strategy}

\begin{figure}[t]
  \centering
   \includegraphics[width=1.0\linewidth]{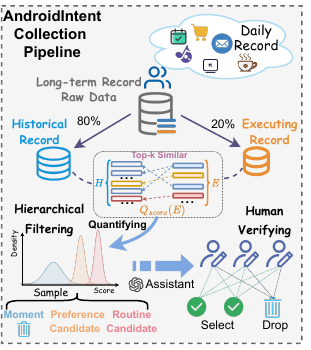}
   \caption{Overview of AndroidIntent collection pipeline. We employ a two-stage filtering-verification, integrating objective criteria with subjective judgment to hierarchically annotate user intent from long-horizon records.}
   \label{fig:pipeline}
\end{figure}

\paragraph{Analysis Strategy.}
Developing personalized agents should first validate stable, modelable personal patterns to establish an existence proof and statistical bounds.
We conceptually distinguish one-off moment intents from recurring preferences and routines based on their occurrence patterns within $\mathcal{H}$, where higher frequency and consistency indicate stable preferences to routines.
To translate these concepts into a practical filtering and validation strategy, we tend to adopt a simple and robust approach that avoids inductive bias and over-pruning. Consequently, we focus on simple and well-established measures of intent semantic similarity and user state distribution.

\noindent \textbf{Intent Semantic Similarity.} 
In analyzing semantic similarity, we employ Qwen3-Embedding \cite{zhang2025qwen3} as our embedding model to measure the semantic similarity between $\mathcal{H}$ and $E$.
For record $R_e \in E$, we compute its similarity with all intents in the historical record $\mathcal{H}$.
Specifically, to prioritize intents with high recurrence, we focus on the density of similar historical instances rather than isolated occurrences.  We retain the top-$k$ most similar records $\mathcal{H}_{k}$ and compute their average similarity score. As a result, intents that are both semantically closer and more frequent in history receive higher scores.
\begin{equation}
\small
\mathcal{S}_{cos}(R_e) = \frac{1}{k} \sum_{\mathcal{H}_{k}}
\text{cos}\big(\text{emb}(I_e), \text{emb}(I_k)\big),
\end{equation}
where $I_e \in R_e$, $I_k \in \mathcal{H}_{k}$, $emb(\cdot)$ denotes the embedding function, and $cos(\cdot)$ represents the cosine similarity between two embeddings.

\begin{figure}[t]
  \centering
   \includegraphics[width=1.0\linewidth]{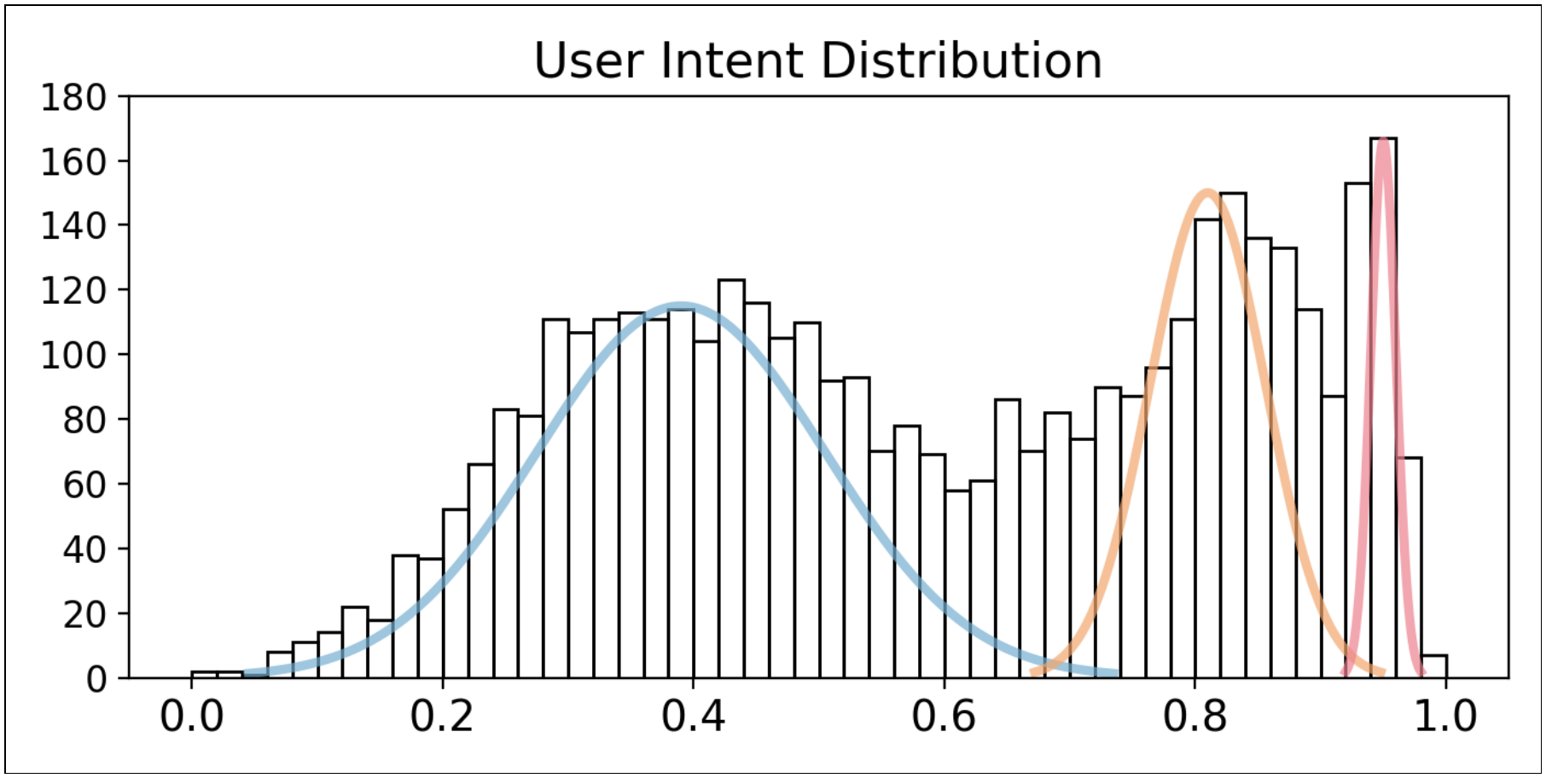}
   \caption{Visualization of the user's intents distribution by aggregating all executing records across users. At a sufficient scale, the intent statistics exhibit three approximately Gaussian-like distributions.}
   \label{fig:analysis}
\end{figure}

\begin{figure*}[t]
  \centering
   \includegraphics[width=1.0\linewidth]{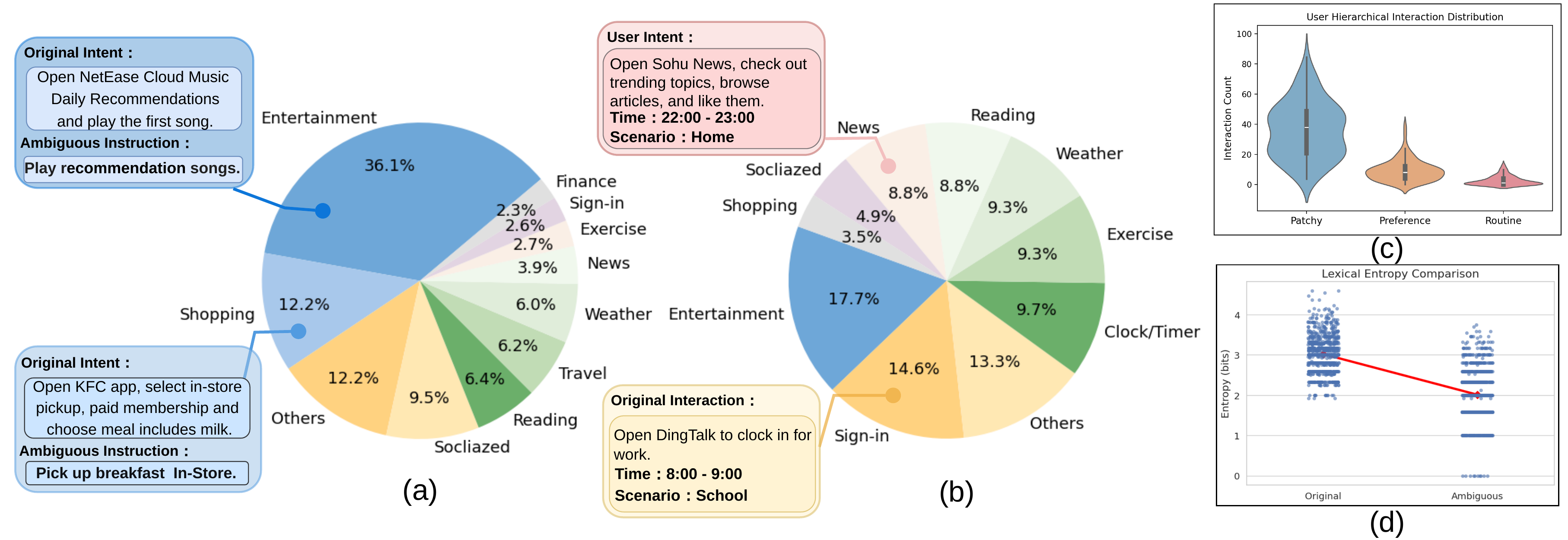}
   \caption{Statistics of AndroidIntent. (a) and (b) show the data distributions for preference and proactive behaviors, respectively; (c) presents the distribution of user instruction types; and (d) illustrates the information volume distribution of vague instructions.}
   \label{fig:statistic}
\end{figure*}

\noindent \textbf{User State Offset.}
In analyzing user state, time $T$ and scenario $S$ serve as key components. We quantify their distribution by analyzing the offsets of $\mathcal{H}_k$ relative to $R_e$ to determine whether similar interactions have been frequently executed under similar states.
Temporal offset entropy $\Delta H_t$ is computed as the normalized entropy over hour-level temporal offsets between $\mathcal{H}_k$ and $R_e$, reflecting the degree of temporal consistency.
Scenario offset entropy $\Delta H_s$ is defined as the normalized entropy over scenario category offsets between $\mathcal{H}_k$ and $R_e$, indicating whether intents are scenario-specific. This can be formulated as:

\begin{equation}
\small
\begin{cases}
\Delta H_t(R_e)= -\frac{\sum\limits
p(\Delta h)\log_2 p(\Delta h)}{
\log_2 hour}
\\[6pt]
\Delta H_s(R_e)= -\frac{\sum\limits p(\Delta s)\log_2 p(\Delta s)
}{\log_2 scene},
\end{cases}
\end{equation}
where $\Delta h$ and $\Delta s$ means the offsets of each $H_k$ to $R_e$, and $hour,scene$ means the total hours and scenarios for normalization. Lower entropy values indicate more stable user states in historical interaction, which are more suitable for proactive.

\paragraph{Quantifying and Filtering.}
Finally, following Eq.~\ref{equ:total}, we compute the quantified score $Q_{score}$ for each executing record in $E$ across all users. As shown in Figure~\ref{fig:analysis}, the resulting distributions over large-scale user data exhibit three hierarchical approximately Gaussian modes, which naturally correspond to moment, preference, and routine intents, showing the existence of modelable patterns for personalized agent at a global mining view.
This distributional structure enables a fast filtering of large-scale data to identify preference and routine candidates, improving both the objectivity and scalability of the annotation.

\begin{equation}
\label{equ:total}
\small
Q_{score}(R_e) = \mathcal{S}_{cos} + \Delta H_t + \Delta H_s
\end{equation}

\subsection{Human Verifying Strategy}
To ensure the final dataset remains unbiased by the filtering process and strictly aligns with human feelings, we perform Human Verifying after filtering.
Following quantification, we slightly expand the sampling range of the Gaussian distribution to retain a broader set of samples in overlapping regions, further mitigating potential overfitting to specific filtering rules.
During annotation, annotators further compare each preference and routine candidate with the user’s historical records $\mathcal{H}$ to re-verify and re-annotate the intent types.
To support the construction of vague preference instructions, we additionally use GPT to generate a diverse set of alternative instructions that deliberately omit potential user preferences. Annotators then select the instruction that best matches the user’s likely intent as the final vague instruction.
All candidates undergo multiple rounds of cross-validation by independent annotators to ensure data quality. 

\subsection{AndroidIntent Statistic}
As shown in Figure \ref{fig:statistic}, we present the overall distribution of AndroidIntent. 
(a) This subset contains \textbf{775} annotated user interactions and \textbf{7,915} GUI actions within 130 different daily apps, with entertainment-related activities accounting for roughly one-third of the data and shopping behaviors comprising 12.2\%. reflecting users' preferences for browsing videos across multiple platforms or purchasing goods on their preferred platforms.
(b) This subset includes \textbf{215} annotated user interactions within 60 different daily apps, among which 14.6\% correspond to sign-in operations. Other proactive intent, such as alarm setting, weather checking, and workout tracking, each account for approximately 9\% and exhibit strong temporal and semantic consistency.
(c) The figure illustrates the distribution of interaction types per user. Moment intents dominate for most users, while each user exhibits, on average, around 10 preference interactions and 3–5 routine interactions. Notably, some users show no clearly identifiable routines.
(d) We compare the lexical entropy of original and vague instructions using a word-level metric. The results show a clear reduction in entropy, indicating that vague instructions tend to be more concise and linguistically simplified.

\begin{table}[t]
\caption{Compare the scale and diversity of AndroidIntent with other recent GUI benchmarks.}
\label{tab:scale}
\centering
\footnotesize
\setlength{\tabcolsep}{2pt}
\begin{tabular}{c |c c | c}
\hline
\bf Benchmark & \bf Episode & \bf Apps & \bf History Records \\
\hline
AndroidWorld &	116 & 20 & - \\
SPA-bench & 340 & 58 & - \\
IFRAgent & 630 & 16	& 9users/7days/15records \\
\hline
\rowcolor{blue!8}
\textbf{Ours} & \textbf{775+215} & \textbf{190} & \textbf{91users/60days/200records}\\
\hline
\end{tabular}
\end{table}

As shown in the Table \ref{tab:scale}, AndroidIntent achieves a competitive scale and substantially improved diversity compared to recent GUI benchmarks. It contains 775+215 episodes spanning 190 applications, significantly exceeding prior datasets in both task coverage and application breadth. In addition, the collected 91 users over 60 days with 200 interaction records provide richer user-centric signals, which are typically limited in user coverage and temporal depth. Overall, these characteristics enable AndroidIntent to support more comprehensive evaluation of user preferences and behavioral routines in realistic GUI interaction scenarios.

Currently, AndroidIntent focuses on Android environments and provides a benchmark for mobile GUI interaction. However, our data filtering and annotation pipeline is applicable across platforms and languages, as it only relies on general GUI interaction primitives such as action trajectories, user interaction states, and user interaction intents. These characteristics are shared across different environments (e.g., web and desktop operating systems), where interactions and user states can be consistently recorded, analyzed, and filtered. As a result, our methodology can be naturally extended to other GUI settings with minimal adaptation. We believe that exploring approaches that can both preserve user privacy and enable scalable collection of user-centric GUI interaction data is a key direction for improving personalized agents.

\section{HIM-Agent}
To support PersonalAlign, agent memory should generalize stable representations to exclude one-off moments while separating preferences and routines, and continuously evolve to stay aligned with user intents. As shown in Figure \ref{fig:method}, we introduce HIM-Agent, a foundational and inspirational personal agent memory that enables GUI agents to rapidly leverage long-term records as context for personalization without interfering with original execution. We construct a streaming update memory and hierarchically organize memory prototypes into Preference Intent Memory and Routine Intent Memory through the execution-based and state-based filter to enable hierarchical intent alignment. 

\subsection{Streaming Aggregation Module}
Raw low-level GUI interaction records are inherently fragmented and noisy. Trivial operating based on these raw record leads to long-tail effects and memory drift, making it difficult to maintain stable personalized representations over time.
To address this challenge, we propose Streaming Aggregation Module that reframes personalization memory from a static log-based storage to a continually evolving representation. Rather than operating in individual records, we maintain Record Prototypes $P_i$ as the fundamental memory units that synthesize similar records into a cohesive whole:
\begin{equation}
\small
P_i = \{R_h | \mathcal{S}_{consist}(R_h, P_i) > \theta \},
\end{equation}
where $R_h \in H$ represents incoming historical records, $\mathcal{S}_{consist}$ measures the consistency between records and prototypes. Based on MicroCluster in stream mining \cite{aggarwal2003framework}, our module incrementally aggregates records at a daily-granularity, enabling an evolving personal memory.

\subsection{Execution-based Preference Filter}
GUI agent memory differs from chat-based memory in that each user interaction includes an execution trajectory rather than purely semantic content. In the Execution-based Filter, we compute $\mathcal{S}_{consist}$ by jointly modeling semantic intent similarity and action trajectory consistency for more comprehensive aggregation for GUI interaction.

\begin{figure}[t]
  \centering
   \includegraphics[width=1.0\linewidth]{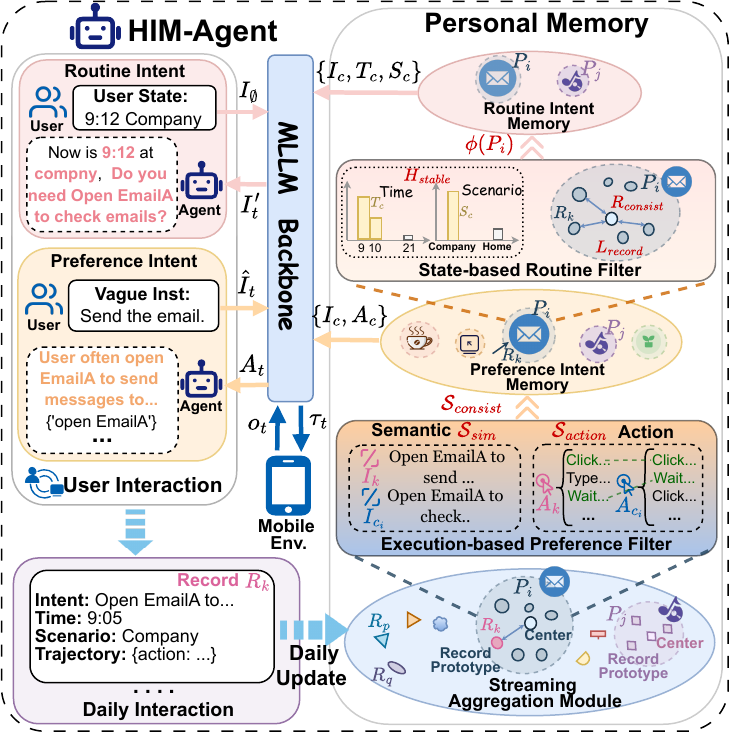}
   \caption{Overview of HIM-Agent. The Streaming Aggregation Module updates user records daily, and the aggregated prototypes are hierarchically organized to support personalized preference and routine intent.}
   \label{fig:method}
\end{figure}

For semantic similarity $\mathcal{S}_{sim}$, we combine dense embedding cosine similarity $\mathcal{S}_{cos}$ with sparse Jaccard $\mathcal{S}_{Jac}$, which computes the overlap ratio of shared words between instructions, to robustly measure semantic similarity. Since GUI instructions are often short and entity-heavy (e.g., app names, items), this may lead to distortions in $\mathcal{S}_{cos}$.

For action consistency $\mathcal{S}_{action}$, we employ Dynamic Time Warping (DTW) to measure the similarity between trajectories that have temporal structure. DTW computes an optimal alignment path $\pi$ by minimizing the cumulative distance between aligned action steps. 
The execution-based preference filter can be formulated as:
\begin{equation}
\small
\begin{cases}
\mathcal{S}_{sim}(I_h, I_{c_i}) = \mathcal{S}_{cos} + \mathcal{S}_{Jac} \\
\mathcal{S}_{action}(A_h, A_{c_i}) = \min\limits_{\pi} \sum_{(i,j) \in \pi} d(a_i, b_j) \\
\mathcal{S}_{consist}(R_h, P_i) = \mathcal{S}_{sim}+ \mathcal{S}_{action},
\end{cases}
\end{equation}
where $I_h, A_h$ denote the intent and action of $R_h$, while $I_{c_i}, A_{c_i}$ represent center intent and action of prototype $P_i$, which are updated daily by selecting the instructions and actions with the minimum average distance to all other records assigned to $P_i$. The pairwise distance $d(a_i,b_j)$ is computed based on the GUI action success rate (SR) \cite{rawles2023androidinthewild, lu2025guiodyssey}, which evaluates whether two actions are the same action.

After filtering, each Record Prototype $P_i$ provides a stable representation and is stored in Preference Intent Memory. When HIM-Agent needs to infer user preferences, the corresponding prototype's center intent $I_c$ and action $A_c$ will be provided.

\begin{table}[t]
\caption{Impact of instruction-induced degradation on GUI agents. Various agents are evaluated under both complete and vague instructions, with results under vague instructions shown in the gray line.}
\centering
\footnotesize
\label{tab:preference}
\begin{tabular}{c | l l l}
\hline
\bf Model & \bf Type & \bf SSR & \bf CSR $\uparrow$ \\ 
\hline
\rowcolor[HTML]{DAE8FC}
\multicolumn{4}{c}{\textit{Closed-sourced GUI Agents}} \\
\hline
\multirow{2}{*}{GPT-5.1} & 51.2 & 26.4 & 52.3  \\
& \cellcolor{gray!20}49.3{\scriptsize\textcolor{red}{$\downarrow$3.7\%}}
& \cellcolor{gray!20}20.3{\scriptsize\textcolor{red}{$\downarrow$23.1\%}}
& \cellcolor{gray!20}22.9{\scriptsize\textcolor{red}{$\downarrow$56.2\%}} \\
\hline
\multirow{2}{*}{GLM-4.5v} & 51.0 & 27.4 & 54.5 \\
& \cellcolor{gray!20}50.5{\scriptsize\textcolor{red}{$\downarrow$0.9\%}}
& \cellcolor{gray!20}19.4{\scriptsize\textcolor{red}{$\downarrow$25.5\%}}
& \cellcolor{gray!20}22.6{\scriptsize\textcolor{red}{$\downarrow$58.5\%}} \\
\hline
\multirow{2}{*}{QwenVL-Max} & 51.9 & 29.8 & 53.3 \\
& \cellcolor{gray!20}51.6{\scriptsize\textcolor{red}{$\downarrow$0.5\%}}
& \cellcolor{gray!20}24.8{\scriptsize\textcolor{red}{$\downarrow$16.9\%}}
& \cellcolor{gray!20}27.3{\scriptsize\textcolor{red}{$\downarrow$48.8\%}} \\
\hline
\rowcolor[HTML]{FFE6CC}
\multicolumn{4}{c}{\textit{Open-sourced GUI Agents}} \\
\hline
\multirow{2}{*}{UI-TARS-1.5}
& 49.4 & 23.5 & 38.6 \\
& \cellcolor{gray!20}46.8{\scriptsize\textcolor{red}{$\downarrow$2.6\%}}
& \cellcolor{gray!20}19.9{\scriptsize\textcolor{red}{$\downarrow$15.3\%}}
& \cellcolor{gray!20}14.9{\scriptsize\textcolor{red}{$\downarrow$23.7\%}} \\
\hline
\multirow{2}{*}{GUI-Owl} & 54.2 & 24.9 & 50.4  \\
& \cellcolor{gray!20}53.1{\scriptsize\textcolor{red}{$\downarrow$2.0\%}}
& \cellcolor{gray!20}17.9{\scriptsize\textcolor{red}{$\downarrow$28.1\%}}
& \cellcolor{gray!20}23.7{\scriptsize\textcolor{red}{$\downarrow$53.0\%}} \\
\hline
\multirow{2}{*}{Qwen3-VL} & 52.7 & 26.7 & 52.9   \\
& \cellcolor{gray!20}46.6{\scriptsize\textcolor{red}{$\downarrow$12.0\%}}
& \cellcolor{gray!20}20.6{\scriptsize\textcolor{red}{$\downarrow$22.8\%}}
& \cellcolor{gray!20}26.6{\scriptsize\textcolor{red}{$\downarrow$49.7\%}} 
\\
\hline

\end{tabular}
\end{table}

\begin{table*}[t]
\caption{GUI agent performance in proactive service. Lower False-Alarm rates indicate better proactive accuracy. Values marked in \underline{\textcolor[RGB]{220,20,60}{red}} denote cases of insufficient capability by the agent.}
\label{tab:proactive}
\centering
\footnotesize
\begin{tabular}{l | c c | c c c c}
\hline
\multirow{2}{*}{\bf Model} & \multicolumn{2}{c|}{\bf Intent Alignment} & \multicolumn{4}{c}{\bf Identification Alignment} 
\\
\cline{2-3} \cline{4-7}
& \bf Semantic & \bf Judgment & \bf Precision & \bf Recall & \bf False-Alarm $\downarrow$ & \bf F1-score\\
\hline
\rowcolor[HTML]{DAE8FC}
\multicolumn{7}{c}{\textit{Closed-sourced GUI Agents}} \\
\hline
\textbf{GPT-5.1} & 49.4\% & 32.0\% & \textbf{73.1\%} & 78.6\% & \textbf{62.0\%} & 75.8\%  \\ 
GLM-4.5V & \textbf{52.8\%} & 33.9\% & 68.9\% & 96.7\% & \underline{\textcolor[RGB]{220,20,60}{94.0\%}} & \textbf{80.4\%} \\ 
QwenVL-Max & 52.2\% & \textbf{34.8\%} & 67.4\% & 97.2\% & \underline{\textcolor[RGB]{220,20,60}{98.0\%}} & 67.4\%  \\ 
\hline
\rowcolor[HTML]{FFE6CC}
\multicolumn{7}{c}{\textit{Open-sourced GUI Agents}} \\
\hline
UI-TARS-1.5 & 42.6\% & 19.0\% & 68.7\% & 99.1\% & \underline{\textcolor[RGB]{220,20,60}{97.0\%}} & \textbf{81.1\%} \\
GUI-Owl & 32.6\% & 12.2\% & 79.9\% & \underline{\textcolor[RGB]{220,20,60}{57.2\%}} & 31.0\% & 66.7\% \\
Qwen3-VL & \textbf{45.0\%} & \textbf{23.6\%} & \textbf{69.0\%} & 97.2\% & \underline{\textcolor[RGB]{220,20,60}{94.0\%}} & 80.7\%\\
\hline
\end{tabular}
\end{table*}

\subsection{State-based Routine Filter}
Upon the formation of a stable prototype $P_i$, we introduce State-based Routine Filter to further separate passive preferences from proactive intents. This module jointly considers the frequency of occurrence, the execution coherence, and the consistency of user states of each $P_i$ to determine whether proactive suggestions should be activated.
To achieve this, we define a proactive confidence $\Phi(P_i)$, which is calculated by: the state stability $H_{state}$, which captures the normalized of temporal and scenario entropies within records of prototype; the record length in prototype $L_{record}$, reflecting how frequently the pattern recurs; and the aggregation weight $R_{consist}$, obtained by averaging $\mathcal{S}_{consist}$. Confidence is jointly inferred based on state consistency, execution consistency, and frequency. This is formulated as:
\begin{equation}
\small
\Phi(P_i) = H_{state} + L_{record} + R_{consist},
\end{equation}

If $\Phi(P_i)$ exceeds the proactive confidence boundary, the corresponding prototype is stored in Routine Intent Memory.
When the HIM-Agent needs to determine whether proactive suggestions are required, prototype's center intent $I_c$ and the most frequent state $T_c,S_c$ will be provided.

\section{Experiment}
\subsection{Experimental Setup}

\begin{table}[t]
\caption{Execution performance across various methods under vague instructions. $\dagger$ denotes the baseline.}
\label{tab:pre-compare}
\centering
\footnotesize
\setlength{\tabcolsep}{2pt}
\begin{tabular}{l|c|cc|cc}
\hline
& \multirow{2}{*}
{\makecell[c]{\textbf{Base}\textsuperscript{$\dagger$}\\\textbf{Model}}} &
\multicolumn{2}{c|}{\bf Retrieve-based} & \multicolumn{2}{c}{\bf Generalized-based} \\
\hhline{~|~|--|--|}
& & \bf Recent & \bf Retrieve & \bf LLM-UM & \cellcolor{blue!8}\bf HIM-Agent \\
\hline
Type & 46.6 & 49.4 & 51.0 & 51.2 & \cellcolor{blue!8}\textbf{52.0} \\
SSR & 20.6 & 21.1 & 22.4 & 22.3 & \cellcolor{blue!8}\textbf{24.0} \\
CSR $\uparrow$ & 26.6 & 33.2 & 35.4 & 35.2 & \cellcolor{blue!8}\textbf{42.3} \\
\hline
\end{tabular}
\end{table}

\paragraph{Metrics.}
We evaluate GUI execution using \textbf{Type Accuracy} (Type) and \textbf{Step-wise Success Rate} (SSR) under an offline evaluation protocol, where treated user actions are the golden trajectory. Moreover, we introduce a new \textbf{Cumulative Successful Rate} (CSR) to measure failures on critical steps caused by vague instructions, where missing user-specific information leads to mismatches with the user’s true intent. To approximate the impact of errors on critical steps, we assign a decaying weight to each step along the trajectory, such that earlier errors contribute more heavily to the overall score. CSR thus serves as an intermediate metric that bridges offline evaluation with online performance.

On the other hand, to evaluate the agent's proactive recommendation capability, we consider \textbf{Intent Alignment} and \textbf{Identification Alignment} \cite{luproactive}. We measure the \textbf{Semantic} similarity between generated suggestions and user’s original intent using embedding cosine similarity and edit distance. We also use an LLM-as-\textbf{Judgment} to evaluate intent alignment, where DeepSeek-V3 is employed to mitigate self-bias.
Identification Alignment evaluates proactive appropriateness, we carefully collect 100 negative user states that do not require proactive assistance, and compute \textbf{Precision}, \textbf{Recall}, \textbf{False-Alarm}, and \textbf{F1-score}.

Please also refer to Appendix~\ref{sec:details} for more details about baselines and implementation details.

\begin{table}[t]
\caption{Proactive performance comparison. FA means False-Alarm rate. $\dagger$ denotes the baseline.}
\label{tab:pro-compare}
\centering
\footnotesize
\setlength{\tabcolsep}{2pt}
\begin{tabular}{l|cc|c >{\columncolor{blue!8}}c}
\hline
& \multicolumn{2}{c|}{\bf Retrieve-based} & \multicolumn{2}{c}{\bf Generalized-based} \\
\hhline{~|--|--|}
& \bf Recent\textsuperscript{$\dagger$} & \bf Retrieve & \bf LLM-UM & \cellcolor{blue!8}\bf HIM-Agent \\
\hline
Semantic & 49.4\% & 49.8\% & 49.1\% & \textbf{53.5\%} \\
Judgement & 32.0\% & 32.2\% & 31.6\% & \textbf{36.3\%} \\
\hline
Precision & 70.8\% & 74.2\% & 75.6\% & \textbf{78.1\%}\\
Recall & 78.3\% & 82.0\% & \textbf{82.3\%} & 81.4\% \\
FA $\downarrow$ & 62.0\% & 64.0\% & 57.0\% & \textbf{49.0\%} \\
F1-score  & 75.8\% & 77.9\% & 78.8\% & \textbf{79.7\%} \\
\hline
Token & 4930 & 3089 & 1161(+6518) & \textbf{1605} \\
\hline
\end{tabular}
\end{table}

\begin{table}[t]
\caption{Ablation study of components in Execution-based Preference Filter Module. Dense and Sparse denote embedding and Jaccard similarity.}
\label{tab:ablation_personal}
\centering
\footnotesize
\begin{tabular}{c c c | c c c}
\hline
\multicolumn{3}{c|}{Components} & \multicolumn{3}{c}{Performance} \\
Dense & Sparse & Action & Type & SSR & CSR$\uparrow$ \\
\hline
\rowcolor{gray!20}
$\times$ & $\times$ & $\times$ & 49.4 & 21.1 & 33.2 \\
$\times$ & $\checkmark$ & $\checkmark$ & 50.8 & 22.5 & 35.9 \\
$\checkmark$ & $\times$ & $\checkmark$ & 51.1 & 22.9 & 36.3 \\
$\checkmark$ & $\checkmark$ & $\times$ & 51.4 & 23.3 & 37.3 \\
\rowcolor{blue!8}
$\checkmark$ & $\checkmark$ & $\checkmark$ & \textbf{52.0} & \textbf{24.0} & \textbf{42.3} \\
\hline
\end{tabular}
\end{table}

\begin{figure}[t]
  \centering
   \includegraphics[width=1.0\linewidth]{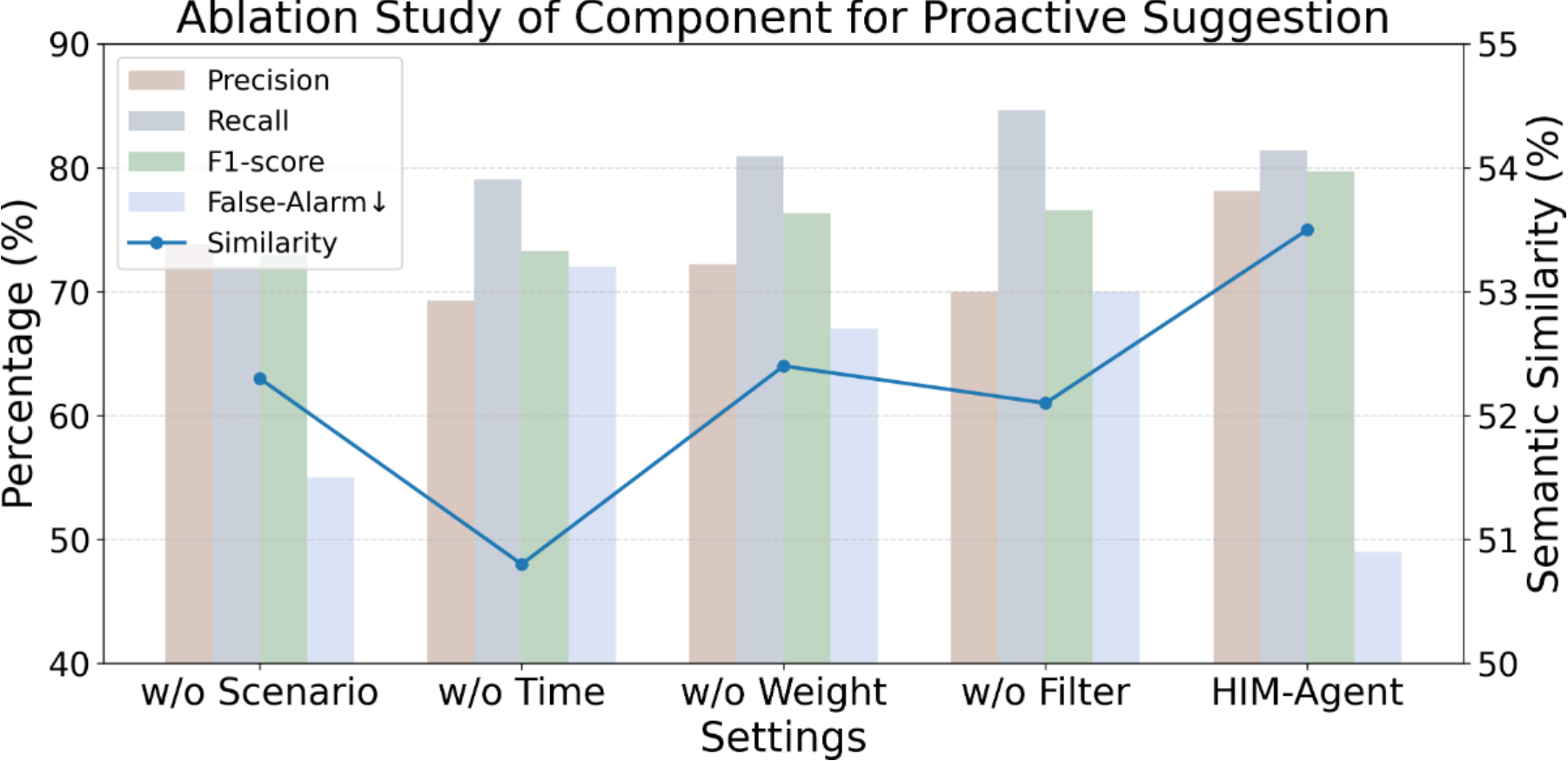}
   \caption{Ablation study of components for proactive performance. Lower of False-Alarm means better align.}
   \label{fig:ablation}
\end{figure}

\begin{figure*}[t]
  \centering
   \includegraphics[width=1.0\linewidth]{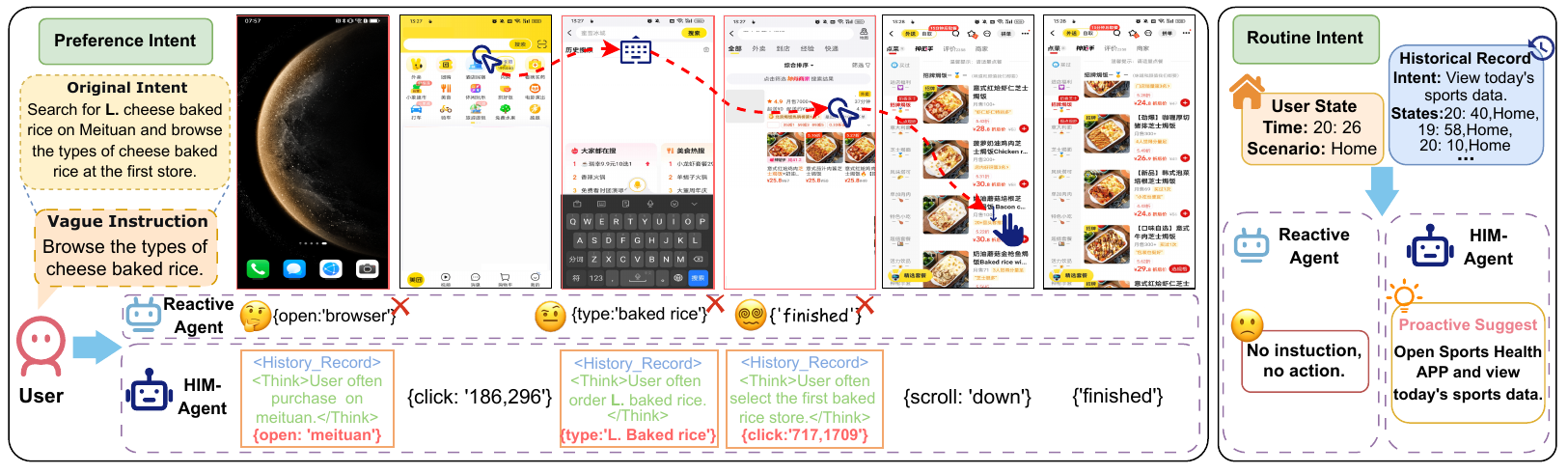}
   \caption{Case study of HIM-Agent. \textbf{Left}: HIM-Agent resolves vague instructions by aligning intent with historical interaction records. \textbf{Right}: HIM-Agent proactively suggests based on historical record and user state.}
   \label{fig:case}
\end{figure*}

\subsection{Experimental Analysis}
\textbf{Vague Instruction Impact on GUI Execution.}
Table~\ref{tab:preference} shows the impact of vague instructions on several outstanding open- and closed-sourced GUI agents. Currently, GUI agents still need to improve their performance across daily instructions and apps since most SSR is around 25-30. Notably, while ambiguity leads to only a 3\% drop in type accuracy, SSR and CSR decrease by approximately 20\% and 45\%. We observe that vague instructions act as \textbf{coarse-grained sub-goals}: although agents can often identify the high-level intended operation, execution fails at a fine-grained level due to the absence of critical personalized preference information. 
For example, lacking explicit requirements, the agent may open incorrect apps, causing execution to deviate significantly from user intent.

\textbf{Challenge in Balancing Proactive Identification.}
Table~\ref{tab:proactive} evaluates the proactive performance of GUI agents based on recent historical records and current user state. Notably, most models struggle to determine when proactive behavior is truly necessary.
Aside from GPT-5.1, current GUI agents generally \textbf{fail to provide effective proactive suggestions}, as they struggle to balance false alarms and recall, often defaulting to overly proactive. These failure cases are highlighted in red and underlined. This reveals a promising research direction: to effectively leverage user records for personalization, agents must develop superior long-term context analysis capabilities.

\textbf{HIM-Agent significantly enhances agent's ability to align implicit intent.}
Since PersonalAlign is a novel paradigm without established methods, we select and compare it against two representative categories of basic approaches: (i) top-down retrieval-based methods, which incorporate recent or relevant historical records as user context for the agent, and (ii) bottom-up inductive methods, which leverage LLMs to summarize user profiles (LLM-UM) \cite{wang2025lettingo}, alongside our HIM-Agent.
As shown in Table~\ref{tab:pre-compare}, we build HIM-Agent based on outstanding open-sourced Qwen3-VL. HIM-Agent achieves the best performance in alleviating the impact of vague instructions, obtaining a CSR score of 42.3.
Furthermore, to objectively evaluate different methods' proactive capability, we conduct experiments on GPT-5.1, which \textbf{only} shows basic balanced proactive ability. As shown in Table~\ref{tab:pro-compare}, our framework achieves superior performance in both Semantic Alignment and Identification Alignment. HIM-Agent helps the agent achieve a better balance between recall 81.4\% and false-alarm 49\% and keep the highest Intent Alignment score 53.3\% and 36.3\%. While LLM-UM introduces extra 6518 tokens consumed when generating user profiles, HIM-Agent remains highly efficient during generalized user modeling.



\textbf{Ablation Study.} Table \ref{tab:ablation_personal} presents the ablation study of components within Execution-based Preference Filter. The first gray line denotes the setting without this filter, under which the streaming aggregation module also can't work, and memory degenerates to individual records without prototypes. The results indicate that all three components contribute to performance gains, and the full module achieves a 9.1\% improvement in CSR.
As shown in Figure \ref{fig:ablation}, we further demonstrate the importance of state-related components in enabling proactive, where both time and scenario play crucial roles. Notably, removing the filter while retaining all prototypes will cause an increase in false alarms to nearly 70\%, which is even higher than the baseline with recent individual records, highlighting the critical role of the state filter for proactive.

\subsection{Case Study}
As shown in Figure \ref{fig:case}, we present case studies comparing HIM-Agent and a reactive agent. In daily usage, user instructions often omit preferences, leading reactive GUI agents to misalign with the user’s true intent. HIM-Agent can infer missing preferences from historical records to correct action execution. By jointly reasoning over records and the current state, HIM-Agent can also proactively assist users, whereas reactive agents remain inactive without explicit instructions.

\section{Conclusion}
We introduce a new critical challenge for agents, PersonalAlign, which requires hierarchical personalization to resolve implicit intents in daily interactions. We introduce AndroidIntent, a new user-centric GUI benchmark curated with a filter-verify strategy, and propose HIM-Agent, a memory framework that enables hierarchical personalization based on long-term user records. Experiments on AndroidIntent show some new challenge and the effectiveness of the HIM-Agent.

\section*{Limitations}
While AndroidIntent and HIM-Agent evaluate how user implicit intent influences GUI agents, several limitations remain. First, the availability of suitable datasets remains limited. Due to the lack of large-scale, publicly available datasets that capture long-term user interaction records, our evaluation is currently restricted to the Fingertip. Our annotation strategy and method cannot yet be broadly validated across diverse real-world datasets. Future work could extend data collection to other GUI environments, such as operating systems and desktop applications.
Second, our method may suffer from a cold-start issue, where insufficient historical interaction data limits the agent’s ability to accurately infer user intent and provide proactive suggestions. Addressing these challenges remains an important and promising direction for future research.

\section*{Ethical considerations}
Since modeling user intent relies on historical interaction data, collecting and utilizing such data may raise privacy potential risks. In our work, we build upon the Fingertip dataset and do not introduce any additional personal information. In the original dataset, all participants were clearly informed about the intended use of the collected data and provided explicit consent by signing a data usage agreement. They were also instructed not to upload any private or sensitive information.
We emphasize that privacy considerations should be treated as a fundamental aspect when developing personalized agents. Future research should explore privacy-preserving solutions, such as on-device deployment, federated learning, or simulated user agents, to ensure that user data remains secure while enabling effective personalization.

\bibliography{main}

\clearpage

\appendix

\makeatletter
\def\thefootnote{\arabic{footnote}}
\makeatother

\section{Experiment Details}
\label{sec:details}

\subsection{Baselines}

In our evaluation on AndroidIntent, we select a diverse set of GUI agent models, including open-source agents such as UI-TARS-7B \cite{qin2025ui}, GUI-Owl-7B \cite{ye2025mobile}, and Qwen-3-VL-8B \cite{bai2025qwen3vl}, alongside several proprietary GUI Agent models, including the GPT-5.1\textsuperscript{1}, GLM-4.5V, and QwenVL-MAX\textsuperscript{2}. We evaluate their GUI execution performance when transitioning from complete to incomplete instructions, and also assess their ability to proactively provide suggestions.

\footnotetext[1]{gpt-5.1-2025-11-13}
\footnotetext[2]{qwen-vl-max(stable)}

\subsection{Implementation details}
All GUI agent execution experiments were conducted on an NVIDIA A100 (40GB) GPU. And we select GPT-5.1 for LLM-UM. During the filtering process, we set $k=10$ for the top-$k$ selection. When computing $Q_{score}$, we use a weighted sum of [1,0.1,0.1] and then normalization. Notably, different weight combinations can yield approximately normal-shaped distributions; we select this configuration to produce clearer decision boundaries, and we also slightly expand the filtering range for moment and preference intent to 0.6 in this setting. Additionally, both $\theta$ in Streaming Aggregation and the proactive boundary threshold in the State-based Routine Filter were both set to 0.6. For CSR, we apply an exponential decay to the length of each trajectory and then perform normalized weighting of SSR. Furthermore, we visualize different hyperparameter combinations of $Q_{score}$ in Figure \ref{fig:qscore}. It can be observed that the three Gaussian-like distributions remain clearly separable across multiple settings, indicating that our filtering strategy is robust to hyperparameter choices. Finally, we select a configuration with relatively clear boundary separation for manual annotation in this paper.

\begin{figure*}[t]
  \centering
   \includegraphics[width=1.0\linewidth]{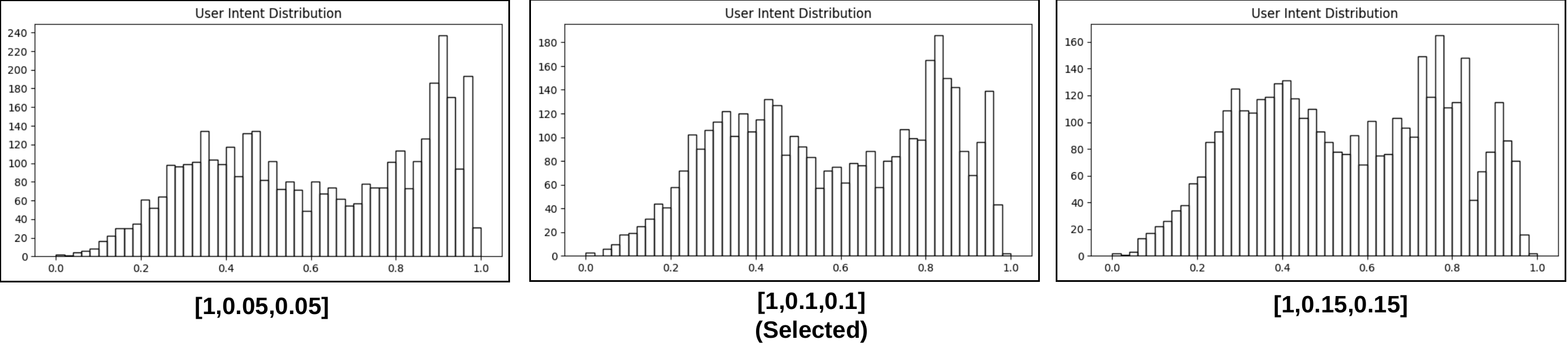}
   \caption{Different combination weights of $Q_{score}$.}
   \label{fig:qscore}
\end{figure*}

\subsection{Online Evaluation}
In this paper, we mainly adopt an offline GUI evaluation setting.
On the other hand, since a user’s intent can often be achieved through multiple valid trajectories, online evaluation is also suitable.  However, unlike simulator-based benchmarks such as AndroidWorld \cite{rawlesandroidworld}, evaluating on AndroidIntent requires connecting to real physical devices via Android Debug Bridge (ADB), as many daily apps cannot run on emulators due to privacy and security restrictions.

Moreover, the evaluation results cannot be automatically verified by Android APIs to determine whether the user intent is successfully completed. Instead, each execution must be manually inspected to assess the agent’s behavior. In addition, real-device evaluation is affected by various factors such as app versions, mobile models, and runtime environments.

Since online evaluation is still not sufficiently stable or scalable under these constraints, we primarily adopt offline evaluation to ensure a more objective and reproducible assessment.

\section{More Analysis}
As shown in Table \ref{tab:proactive}, we observe that most existing GUI agents still struggle to effectively perform proactive behaviors based on user history. To further analyze, we additionally evaluate several models with stronger long-context understanding capabilities. As reported in Table \ref{tab:proactive_other}, Gemini3-Pro also demonstrates relatively strong proactive performance under the same prompt. In contrast, models from the Qwen series (e.g., Qwen-3 and Qwen-Max) tend to behave overly aggressively under the same prompts, frequently triggering proactive actions and thus leading to a high false-alarm rate.
Future research on end-to-end or on-device personal GUI agents should not only focus on execution success rates, but also carefully consider the model’s ability to understand and reason over long-term context, which is crucial for analyzing user record context.

\begin{table*}[t]
\caption{Comparison of more models in proactive capability. }
\label{tab:proactive_other}
\centering
\footnotesize
\begin{tabular}{l | c c | c c c c}
\hline
\multirow{2}{*}{\bf Model} & \multicolumn{2}{c|}{\bf Intent Alignment} & \multicolumn{4}{c}{\bf Identification Alignment} 
\\
\cline{2-3} \cline{4-7}
& \bf Semantic & \bf Judgment & \bf Precision & \bf Recall & \bf False-Alarm $\downarrow$ & \bf F1-score\\
\hline
GPT-5.1 & 49.4\% & 32.0\% & 73.1\% & 78.6\% & 62.0\% & 75.8\%  \\
Gemini3-Pro & 53.5\% & \textbf{37.0\%} & \textbf{74.7\%} & 77.8\% & \textbf{57.0\%} & 76.2\% \\
Qwen-Max & \textbf{53.6\%} & 35.4\% & 68.7\% & \textbf{93.1\%} & \underline{\textcolor[RGB]{220,20,60}{92.0\%}} & \textbf{79.1\%} \\
\hline
Qwen3-8B & 49.3\% & 26.8\% & 69.5\% & 95.4\% & \underline{\textcolor[RGB]{220,20,60}{90.0\%}} & 80.3\% \\
\hline
\end{tabular}
\end{table*}

\section{Personal Agent Settings}
\label{sec:personal-setting}
With long-term user records, we can further envision more intelligent personal agents that provide personalized assistance in everyday life, similar to \textit{Jarvis}, which offers highly customized and proactive services.

\subsection{Personalized Rewriting}
Preference inference can also be implemented via a query rewriting strategy. Specifically, we feed the user’s vague instruction together with historical records into an LLM to perform Personalized Rewriting, which can be formulated as:
\begin{equation}
    I_{re}' \leftarrow f_{\theta}(\hat{I}; \{(I_i, T_i, S_t)\} \in \mathcal{H}).
\end{equation}
where $I_{re}'$ means LLM rewrite instruction. However, this approach introduces an additional inference stage and intermediate variables, thereby breaking the end-to-end nature of GUI agents. More importantly, it only enables personalization at the instruction-level. In contrast, PersonalAlign supports finer-grained personalization at the action-level, providing more direct control over execution behaviors.

\subsection{Proactive Triggering}
From a user-centric perspective, proactive agents can also be designed in a trigger-based paradigm. Specifically, by summarizing historical interaction records, the agent can learn a set of trigger states defined by specific temporal and contextual conditions under which recommendations should be initiated. This can be formulated as:
\begin{equation}
    ({T_t', S_t'}) \leftarrow f_{\theta}(I_\emptyset ;\{(I_i, T_i, S_i)\} \in \mathcal{H}).
\end{equation}

In this setting, the agent does not need to continuously reason over the current state at every moment; instead, it activates recommendations only when predefined trigger conditions are met.

However, evaluating such a framework requires an online-like environment, where a User Agent  \cite{luproactive} should be constructed to simulate user behavior by triggering predefined states and providing accept or reject feedback. Overall, compared to real-time reasoning, this trigger-based paradigm is more efficient and represents a promising direction for future work.

\subsection{Proactive Executing}
Imagine a future scenario: upon arriving at the office, your phone is already unlocked and waiting at the relevant workspace interface; or when you leave the company, the navigation app automatically opens and sets the route home. This represents a more forward-looking paradigm of interaction, which we refer to as Proactive Executing, which can be formulated as:
\begin{equation}
    A_t \leftarrow f(I_\emptyset; \{(I_i, T_i, S_i)\} \in \mathcal{H}).
\end{equation}

Such proactive agents constitute a more advanced instantiation of HIM-Agent. This paradigm goes beyond reactive or recommendation-based behaviors by anticipating user needs and executing actions autonomously. However, realizing and evaluating such capabilities requires an online evaluation environment, along with a simulated user agent capable of modeling realistic acceptance or rejection behaviors.

As GUI Agents continue to advance in capability, we envision that stronger personalization, contextual awareness Proactive Executing Agents will become achievable. This represents a promising direction toward truly intelligent and personalized human–agent interaction.

At present, such a setting remains challenging to realize in practice. Therefore, in PersonalAlign, we restrict proactive behavior to the instruction level, where we evaluate the agent’s ability to predict user intent rather than executing actions autonomously. 
On the other hand, proactive decision-making can also be informed by richer signals; in AndroidIntent, we primarily focus on the user's interaction time and scenario, leaving the incorporation of additional factors to future work.


\section{Annotator Requirements}
\label{sec:guidance}
Here, we present the requirements provided to the annotators. To better support them in verifying candidate results, we also design an interactive interface that facilitates faster and more efficient annotation.

We recruited a group of student annotators living in China to annotate and provided compensation that aligns with local living wage standards. The annotation workload during the verifying stage was intentionally kept lightweight, as we designed a user-friendly interface to assist the annotation process and reduce cognitive burden. This design ensured that the workload remained reasonable while maintaining annotation quality.

\begin{lstlisting}[style=requirements]
You are provided with a ranked list of historical records, ordered by their relevance to the current intent. For each record, you can view the interaction time, scenario, and action. You should determine which type of user behavior the current intent belongs to, based on the user's history.

We define three types of user intents:

1. Moment Intent

If the current intent rarely appears or does not appear in the history, and you believe that a complete and explicit description is required for the agent to execute it correctly.

2. Preference Intent

If similar intent have appeared multiple times in history, and you believe the user has performed this action repeatedly, such that only minimal information is needed and the remaining details can be inferred from past interactions.

After selecting this option, you are required to choose one personalized instruction from the provided vague instruction candidates that you think the user would most likely give. If none are suitable, you may also write your own possible instructions.

3. Routine Intent

If the intent has appeared many times in nearly identical forms, and you believe the user has performed it frequently enough that the agent can infer the intent directly from historical patterns, especially from the time and scenario, without any additional information.

Except for the preference intent, no additional instruction selection is required.

After finishing the annotation for one intent, you can click `next`, and the interface will automatically update the intent and user history.

If you are uncertain about a case, you may also mark it as uncertain for further review.
\end{lstlisting}

\section{Prompt for Agents}
\label{sec:prompt}
\subsection{Prompt for LLM-UM}
Prompt for the preference interaction summary:
\begin{lstlisting}[style=promptbox]
You will serve as an assistant to help me summarize a user's preferences based on his/her long-term GUI usage records. I will provide you with the user's brief profile amd long-term interaction history. From this history, you need to extract the user's preferences regarding apps or items.

## Requirements:

1. Please provide your output in JSON format, following this structure:
{
 "summaries": [
  {
    "reasoning": "Briefly explain your reasoning for the preference",
    "preference": "A concise summary of what types of apps/items this user is likely to enjoy, e.g. 'User preference for shopping with App A'",
    "confidence": "High/Medium/Low",
    "action": "Summarize one User's usual execution actions trajectory from `action_list`, e.g. ['click(x1, y1)', 'wait()', 'finished()']"
    (if you are unable to determine the preference, please set this value to "None")
},
]
}
2. Ensure that each "preference" is highly concise. It should clearly describe one app + one specific type of task/content the user prefers.
3. You must provide multiple preference-reasoning pairs with JSON format in "summaries". Each preference should reference only one app
4. The answer should be in Chinese. However, "action" field must keep the original English types from the `action_list`.
5. Do not provide any text outside of the JSON string.
6. Additional requirement:  
   - If several apps appear interchangeable but one is used significantly more often for a specific task, treat that as a preference.  
   - Avoid fabricating preferences when evidence is weak; use 'None' when uncertain.
   
## User Profile:
{profile}

## User History:
{previous_intents}
\end{lstlisting}

\subsection{Prompt for Proactive}
\begin{lstlisting}[style=promptbox]
You are skilled at analyzing user history. You are given summarized user's daily routine. Your task is to determine whether the current user state requires proactive suggestions. 

## Note
- You are given summarized user routines, each describing a frequent user behavior with its intent, time/scenario distribution, and frequency.
- If neither the time nor the scenario shows a sufficiently strong match with any summarized routine, output False.
- If you decide that a recommendation is needed, output a suitable user instruction for the GUI agent to execute.
- Express the user's intent unvaguely in one Chinese sentence, ensuring the expression faithfully reflects the user's original intent without adding any extra description.
- Do not output any explanation, including time and scenario information. Imitate the user's intent and output only one proactive suggestion.

## Input
Time: {time}
Scenario: {scenario}
Summarized_routine: {generalized_routine}

The user's intent:
\end{lstlisting}

\section{AI Assistant}
We used AI assistants only for minor writing assistance during manuscript preparation.

\end{document}